\documentclass[letterpaper, 10 pt, conference]{ieeeconf}  
\pdfoutput=1

\IEEEoverridecommandlockouts                              

\usepackage[pdftex]{graphics} 
\usepackage{epstopdf}
\usepackage{epsfig} 
\usepackage{hyperref}
\usepackage{amsmath}
\usepackage{amssymb}  
\usepackage{mathtools}
\usepackage{units}
\usepackage{cleveref}
\usepackage{color}
\usepackage{colortbl}

\def\b#1{\boldsymbol{#1}}
\def\Rot{\Phi}
\def\rot{\b{\varphi}}
\def\s{\sin}
\def\c{\cos}

\definecolor{myGreen}{rgb}{0.6, 1.0, 0.6}
\definecolor{myRed}{rgb}{1.0, 0.6, 0.6}
\definecolor{myBlue}{rgb}{0.6, 0.6, 1.0}

\title{\LARGE \bf
A Primer on the Differential Calculus of 3D Orientations
}

\author{Michael Bloesch, Hannes Sommer, Tristan Laidlow, Michael Burri, Gabriel Nuetzi, P\'eter Fankhauser, \\ Dario Bellicoso, Christian Gehring, Stefan Leutenegger, Marco Hutter, Roland Siegwart}

\begin{document}

\maketitle
\thispagestyle{empty}
\pagestyle{empty}
\begin{abstract}
The proper handling of 3D orientations is a central element in many optimization problems in engineering.
Unfortunately many researchers and engineers struggle with the formulation of such problems and often fall back to suboptimal solutions.
The existence of many different conventions further complicates this issue, especially when interfacing multiple differing implementations.
This document discusses an alternative approach which makes use of a more abstract notion of 3D orientations.
The relative orientation between two coordinate systems is primarily identified by the coordinate mapping it induces.
This is combined with the standard exponential map in order to introduce representation-independent and minimal differentials, which are very convenient in optimization based methods.
\end{abstract}

\section{Introduction}
The primary goal of this document is to derive and summarize the most important identities for handling 3D orientations.
It can readily be used as a look-up document (general identities are green (\Cref{sec:indide}), implementation dependent identities are red (\Cref{sec:quaImp})).
In a compact theoretical section all equations are derived together with some insights into their mathematical background (\Cref{sec:the}).
We believe however, that the best way to understand these concepts is to apply the presented findings on an actual system.
To this end, we discuss the modeling of an Inertial Measurement Unit (IMU) driven kinematic model (\Cref{sec:modexa}).
Furthermore, we provide the most important proofs and derivations in order to provide some additional insights and examples.
Similar elaborations on the topic exist in \cite{drummond,eade,barfoot_ser15}.

An understanding of kinematics (including the concept of coordinate systems) is a prerequisite for understanding this document.
The corresponding conventions and notations are summarized in \Cref{sec:not}.
To completely follow the theoretical sections some higher mathematical concepts are necessary. 

\section{Vectors and Coordinate Systems Notation} \label{sec:not}
In this document coordinate systems are denoted by calligraphic capital letters, e.g. $\mathcal{A}$, and coordinate tuples are represented by bold lower case letters, e.g.  ${}_\mathcal{A}\b{r}_{\mathcal{B}\mathcal{C}}$.
The left-hand subscript of a coordinate tuple indicates the coordinate system the vector is represented in, while the right-hand subscripts indicate the 3D points related to start and end points.
For instance, the term ${}_\mathcal{A}\b{r}_{\mathcal{B}\mathcal{C}}$ denotes the coordinates of a vector $\vec{\b{r}}_{\mathcal{B}\mathcal{C}}$ (denoted with an arrow) in the Euclidean space $\mathbb{E}^3$ from point $\mathcal{B}$ to point $\mathcal{C}$, represented in the coordinate system $\mathcal{A}$.
By abuse of notation, we denote the origin associated with a specific coordinate system by the same symbol.
Furthermore, the term $\Rot_{\mathcal{B}\mathcal{A}} \in SO(3)$ is employed for representing the relative orientation of a coordinate system $\mathcal{B}$ w.r.t. a coordinate system $\mathcal{A}$.
Its definition is coupled to the (distance preserving) mapping of coordinate tuples and we employ the notation ${}_\mathcal{B}\b{r}_{\mathcal{B}\mathcal{C}} = \Rot_{\mathcal{B}\mathcal{A}} \left({}_\mathcal{A}\b{r}_{\mathcal{B}\mathcal{C}}\right)$.
We define the mapping $\b{C}: SO(3) \to \mathbb{R}^{3 \times 3}$ such that $\Rot(\b{r}) \triangleq \b{C}(\Rot) \b{r}$ (corresponding to the rotation matrix).
A more complete overview of coordinate systems and rotations is given in \cite{barfoot_ser15}.

Furthermore, the vectors $\vec{\b{v}}_{\mathcal{B}}$ and $\vec{\b{a}}_{\mathcal{B}}$ denote the absolute (w.r.t. an inertial coordinate system) velocity and acceleration of the point $\mathcal{B}$.
The vector $\vec{\b{\omega}}_{\mathcal{A}\mathcal{B}}$ denotes the relative angular velocity of the coordinate system $\mathcal{B}$ w.r.t. the coordinate system $\mathcal{A}$.
The skew symmetric matrix of a coordinate tuple $\b{v} \in \mathbb{R}^3$ is denoted as $\b{v}^\times \in \mathbb{R}^{3 \times 3}$ and has the property $\b{v}^\times \b{r} = \b{v} \times \b{r} \ \forall \b{r} \in \mathbb{R}^3$, where $\times$ denotes the Euclidean cross-product.
The term $\b{v}^\times$ fulfills the following identities ($\b{I} \in \mathbb{R}^{3 \times 3}$ is the identity matrix):
\begin{align}
  (\b{v}^\times)^T &= -\b{v}^\times, \\
  (\b{v}^\times)^2 &= \b{v}\b{v}^T - \b{v}^T\b{v} \b{I}, \\
  (\b{C}(\Rot)\b{v})^\times &= \b{C}(\Rot) \b{v}^\times \b{C}(\Rot)^T.
\end{align}

\begin{figure}
\centering
\includegraphics[width = \columnwidth]{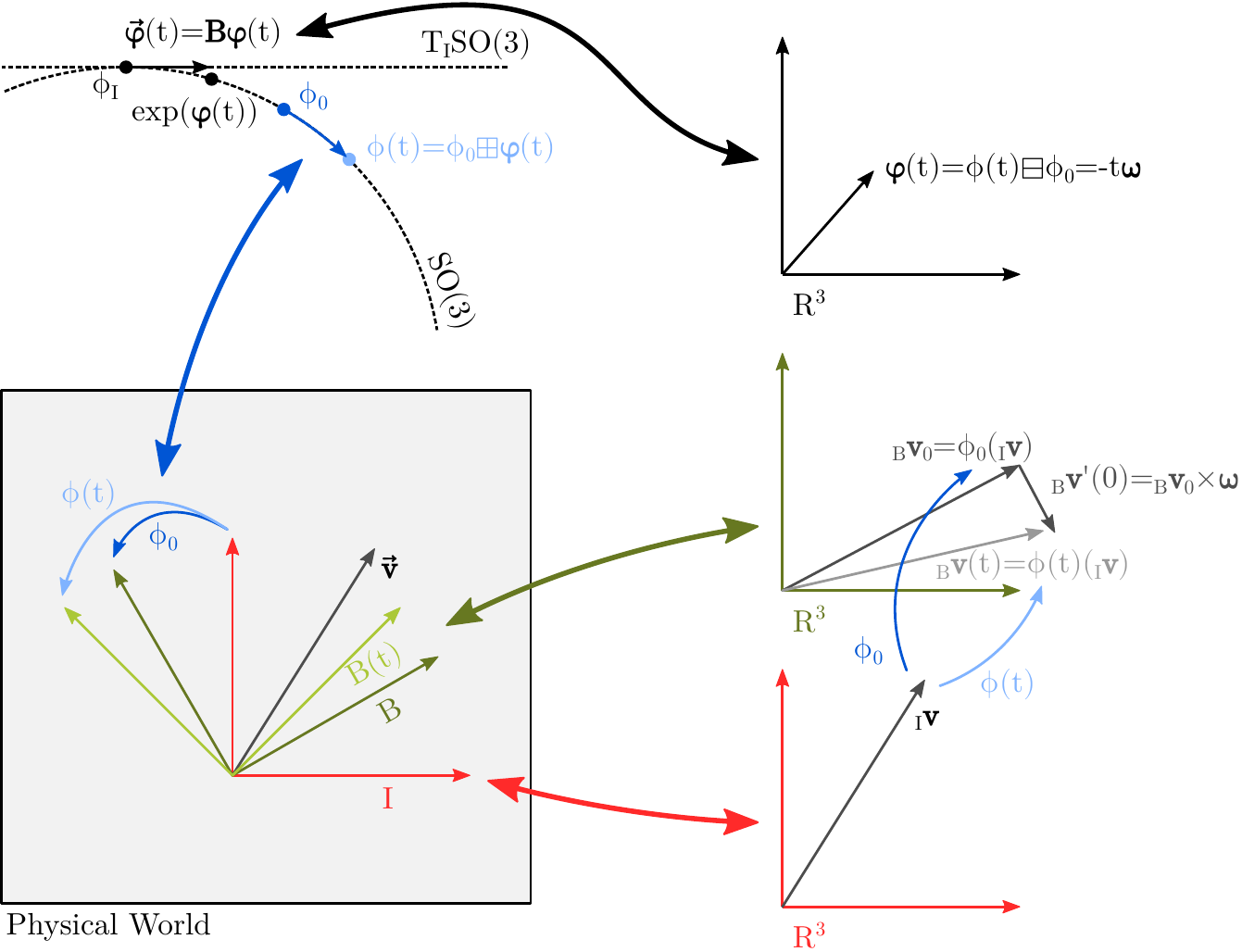}
\caption{This figure depicts various quantities in a setup where a coordinate system $\mathcal{B}$ is rotated by a constant rotational velocity $\b{\omega}$ w.r.t. an inertial coordinate system $\mathcal{I}$. Using coordinate systems, physical vectors can be represented through the corresponding coordinate tuples.
Orientations between coordinate systems can be defined by the mapping they induce on the coordinate tuples.
They are elements of $SO(3)$.
Differences and derivatives of orientations can be represented in the tangential space $T_{\Rot_I}SO(3)$, which can be associated with $\mathbb{R}^3$ by means of a basis $\b{B}$.}
\label{fig:orialg}
\end{figure}

\section{Theory} \label{sec:the}
The following contemplations are independent of the choice of parametrization for 3D orientations.
As will follow in the next definition, 3D orientations are first only thought of as mapping.

Given a 3D rigid body with attached body-fixed coordinate system $\mathcal{B}$, its orientation $\Rot_{\mathcal{B}\mathcal{A}}$ w.r.t. a reference coordinate system $\mathcal{A}$ can be defined as the mapping of coordinates of any fixed vector $\vec{\b{r}}$ from $\mathcal{A}$ to $\mathcal{B}$, that is,
\begin{align}
  {}_\mathcal{B}\b{r} = \Rot_{\mathcal{B}\mathcal{A}}({}_\mathcal{A}\b{r}).
\end{align}
Together with the concatenation operation, orientations form a Lie group known as the special orthogonal group $SO(3)$.
The concatenation $\circ: SO(3) \times SO(3) \to SO(3)$ comes with the following (defining) identity:
\begin{align}
  (\Rot_\mathcal{CB} \circ \Rot_\mathcal{BA})({}_\mathcal{A}\b{r}) \triangleq \Rot_\mathcal{CB} (\Rot_\mathcal{BA}({}_\mathcal{A}\b{r})). \label{eq:condef}
\end{align}
There also exists an identity element $\Rot_I$ and an inverse $\Rot^{-1}$ such that
\begin{align}
  \Rot_I \circ \Rot_\mathcal{BA} &= \Rot_\mathcal{BA} \circ \Rot_I = \Rot_\mathcal{BA}, \\
  \Rot_\mathcal{BA}^{-1} \circ \Rot_\mathcal{BA} &= \Rot_\mathcal{BA} \circ \Rot_\mathcal{BA}^{-1} = \Rot_I.
\end{align}

The Lie group $SO(3)$ is not a vector space, has no addition operation, and consequently no subtraction either.
This poses an issue if using orientations in filtering or optimization frameworks, which strongly rely on small differences and gradients (e.g. for linearization).
Fortunately, since $SO(3)$ is a Lie group, there exists an exponential map $Exp: T_{\Rot_I}SO(3) \to SO(3)$ relating $SO(3)$ to its Lie algebra $T_{\Rot_I}SO(3)$.
The later coincides with the tangent space at the identity element, which is isomorphic to $\mathbb{R}^3$.
The exponential map is smooth and fulfills the following (uniquely) defining identities $\forall t,s \in \mathbb{R}, \ \forall \vec{\b{\varphi}} \in T_{\Rot_I}SO(3)$:
\begin{align}
  Exp((t+s) \vec{\b{\varphi}}) &= Exp(t \vec{\b{\varphi}}) \circ Exp(s \vec{\b{\varphi}}), \label{eq:expCon1} \\
  \nicefrac[]{d}{dt} (Exp(t \vec{\b{\varphi}}))|_{t=0} &= \vec{\b{\varphi}}. \label{eq:expCon2}
\end{align}

Elements on $T_{\Rot_I}SO(3)$ are abstract vectors and are not very suitable for actual computations.
By choosing a basis $\b{B} = [\vec{\b{\varphi}}_1, \vec{\b{\varphi}}_2, \vec{\b{\varphi}}_3]$ the map can be extended to $\mathbb{R}^3$.
We define the exponential $\exp: \mathbb{R}^3 \to SO(3)$ of a coordinate tuple $\rot = (\varphi_1,\varphi_2,\varphi_3) \in \mathbb{R}^3$ by 
\begin{align}
\exp(\rot) := Exp(\vec{\b{\varphi}}_1 \varphi_1 + \vec{\b{\varphi}}_2 \varphi_2 + \vec{\b{\varphi}}_3 \varphi_3). \label{eq:expmapdef}
\end{align}
There is a certain degree of freedom in the selection of the basis $[\vec{\b{\varphi}}_1, \vec{\b{\varphi}}_2, \vec{\b{\varphi}}_3]$.
We choose the basis vectors $\vec{\b{\varphi}}_i$ such that $\forall i \in \{1,2,3\}, \forall \b{v} \in \mathbb{R}^3$:
\begin{align}
  \nicefrac[]{d}{dt} (Exp(t \vec{\b{\varphi}}_i)(\b{v}))|_{t=0} = \b{e}_i \times \b{v}
\end{align}
where $\b{e}_i \in \mathbb{R}^3$ are the standard basis vectors in $\mathbb{R}^3$.
This makes $\exp(\cdot)$ a unique smooth mapping that fulfills $\forall t,s \in \mathbb{R}, \ \forall \rot,\b{v} \in \mathbb{R}^3$:
\begin{align}
  \exp((t+s) \rot) &= \exp(t \rot) \circ \exp(s \rot) \label{eq:exppCon1} \\
 \nicefrac[]{d}{dt} (\exp(t \rot)(\b{v}))|_{t=0} &= \rot \times \b{v} \label{eq:exppCon2}
\end{align}
We will see later, that by using this definition of the exponential $\exp$, its argument $\rot$ can be interpreted as the rotation vector associated with the relative orientation of two coordinate systems.
There exists an open region around 0, the open ball with radius $\pi$ $B_\pi(0) \subset \mathbb{R}^3$, such that the exponential is bijective and its image corresponds to all non-$180^{\circ}$-orientations, $SO(3)^*$.
Thus an inverse exists which is called the logarithm, $\log: SO(3)^* \to B_\pi(0)$.

With this we can construct boxplus and boxminus operations which adopt the function of addition and subtraction \cite{Hertzberg2011}:
\begin{align}
	\boxplus : & SO(3) \times \mathbb{R}^{3} \rightarrow SO(3), \label{eq:boxplus} \\
	& \Rot, \rot \mapsto \exp(\rot) \circ \Rot, \nonumber \\
	\boxminus : & SO(3) \times SO(3) \rightarrow \mathbb{R}^{3}, \label{eq:boxminus} \\
	& \Rot_1, \Rot_2 \mapsto \log(\Rot_1 \circ \Rot_2^{-1}). \nonumber
\end{align}
Similarly to regular addition and subtraction, both operators fulfill the following identities (axioms proposed by \cite{Hertzberg2011}):
\begin{align}
	\Rot \boxplus \b{0} &= \Rot, \\
	(\Rot \boxplus \rot) \boxminus \Rot &= \rot, \\
	\Rot_1 \boxplus (\Rot_2 \boxminus \Rot_1) &= \Rot_2.
\end{align}
This approach distinguishes between actual orientations, which are on $SO(3)$ (Lie group), and differences of orientations which lie on $\mathbb{R}^3$ (Lie algebra, see \Cref{fig:orialg}).
The above operators take care of appropriately transforming the elements into their respective spaces and allow a smooth embedding of rotational quantities in filtering and optimization frameworks.

The definition of differentials involving orientations can be adapted by replacing the regular plus and minus operators by the above boxplus and boxminus operators.
For instance the differential of a mapping $f_1: \mathbb{R} \to SO(3)$ can be defined as:
\begin{align}
\frac{\partial}{\partial x} f_1(x) :=& \lim_{\epsilon \rightarrow 0} \frac{f_1(x+\epsilon) \boxminus f_1(x)}{\epsilon}. \label{eq:der1}
\end{align}
The same can be done for the other case where we have a mapping $f_2: SO(3) \to \mathbb{R}$:
\begin{align}
\frac{\partial}{\partial \Rot} f_2(\Rot) :=& \lim_{\epsilon \rightarrow 0} \begin{bmatrix}
	\frac{f_2(\Rot \boxplus (\b{e}_1 \epsilon)) - f_2(\Rot)}{\epsilon} \\ \frac{f_2(\Rot \boxplus (\b{e}_2 \epsilon)) - f_2(\Rot)}{\epsilon} \\ \frac{f_2(\Rot \boxplus (\b{e}_3 \epsilon)) - f_2(\Rot)}{\epsilon}
	\end{bmatrix}^T. \label{eq:der2}
\end{align}

\section{Implementation-Independent Identities} \label{sec:indide}
Some identities directly follow from the above considerations and are \emph{independent} of the choice of the underlying orientation representation.
By concatenating the exponential and the coordinate mapping we retrieve the well known Rodriguez' formula (see Appendix \ref{ap:rodfor}):

\vspace{2pt}
\noindent
\colorbox{myGreen}{\parbox[c][60pt]{\columnwidth-2\fboxsep}{
\begin{align}
	\b{C}(\rot) &:= \b{C}(\exp(\rot)) \label{eq:rod} \\
    &= \b{I} + \frac{\s(\|\rot\|) \rot^\times}{\|\rot\|} + \frac{(1 - \c(\|\rot\|)) \rot^{\times^2}}{\|\rot\|^2}, \nonumber \\
	\b{C}(\rot) &\approx \b{I} + \rot^\times, \ \ (\|\rot\| \approx 0). \label{eq:rodsmall} 
\end{align}
}}\vspace{2pt}
This shows that the argument of the exponential, $\rot$, can be interpreted as the coordinate tuple of the (\emph{passive}) rotation vector associated with the relative orientation of two coordinate systems.
Thus, if the corresponding coordinate systems are known we can write:
\begin{align}
\Rot_\mathcal{BA} = \exp\left({}_\mathcal{B}\rot_\mathcal{BA}\right) = \exp\left({}_\mathcal{A}\rot_\mathcal{BA}\right). \label{eq:rotvec}
\end{align}
We can also derive the following (adjoint related) identity (see Appendix \ref{ap:adjoint}):

\vspace{2pt}
\noindent
\colorbox{myGreen}{\parbox[c][15pt]{\columnwidth-2\fboxsep}{
\begin{align}
	\exp(\Rot(\rot)) &= \Rot \circ \exp(\rot) \circ \Rot^{-1}.\label{eq:adjoint}
\end{align}
}}\vspace{2pt}

Useful identities can be derived for derivatives involving orientations (see Appendix \ref{app:der}):

\vspace{2pt}
\noindent
\colorbox{myGreen}{\parbox[c][124pt]{\columnwidth-2\fboxsep}{
\begin{align}
	\partial/\partial t \left(\Rot_{\mathcal{B}\mathcal{A}}(t)\right) &= -{}_\mathcal{B}\b{\omega}_{\mathcal{A}\mathcal{B}}(t), \label{eq:dertime}\\
	\partial/\partial \b{r} \left(\Rot(\b{r})\right) &= \b{C}(\Rot), \\
	\partial/\partial \Rot \left(\Rot(\b{r})\right) &= -\left(\Rot(\b{r})\right)^\times, \label{eq:dercm}\\
	\partial/\partial \Rot \left(\Rot^{-1}\right) &= - \b{C}(\Rot)^{T}, \label{eq:derinv}\\
	\partial/\partial \Rot_1 \left(\Rot_1 \circ \Rot_2\right) &= \b{I}, \label{eq:dercon1} \\
	\partial/\partial \Rot_2 \left(\Rot_1 \circ \Rot_2\right) &= \b{C}(\Rot_1), \label{eq:dercon2}\\
	\partial/\partial \rot \left(\exp(\rot)\right) &= \b{\Gamma}(\rot), \label{eq:derexp}\\
	\partial/\partial \Rot \left(\log(\Rot)\right) &= \b{\Gamma}^{-1}(\log(\Rot)\label{eq:derlogmap}).
\end{align}
}}\vspace{2pt}
The derivative of the exponential map is given by the Jacobian $\b{\Gamma}(\rot) \in \mathbb{R}^{3 \times 3}$ which has the following analytical expression:

\vspace{2pt}
\noindent
\colorbox{myGreen}{\parbox[c][55pt]{\columnwidth-2\fboxsep}{
\begin{align}
	\b{\Gamma}(\rot) &= \b{I} + \frac{(1 - \c(\|\rot\|)) \rot^\times}{\|\rot\|^2} + \frac{(\|\rot\| - \s(\|\rot\|)) \rot^{\times^2}}{\|\rot\|^3}, \label{eq:gamma} \\
	\b{\Gamma}(\rot) &\approx \b{I} + 1/2 \rot^\times, \ \ (\|\rot\| \approx 0). 
\end{align}
}}\vspace{2pt}

\section{Quaternion Implementation} \label{sec:quaImp}
The above discussion is completely decoupled from any actual orientation parameterization.
It is valid whether Euler-angles, rotation matrices, quaternions, or other representations are employed.
In the following we provide one possible \emph{implementation} of 3D orientations along with the means to check its correctness.
Here we propose the use of unit quaternions following the \emph{Hamilton} convention \cite{hamilton1844} and we discuss the implementation of the different operations that are required.
For more details on the differences between existing quaternion conventions we refer the reader to \cite{sola2012quaternion}.
A unit quaternion is composed of a real part, $q_0 \in \mathbb{R}$, and an imaginary part, $\check{\b{q}} \in \mathbb{R}^3$, which meet $q_0^2 + \|\check{\b{q}}\|^2 = 1$.
We denote this as $\Rot = (q_0, \check{\b{q}})$.

\subsection{Coordinates Mapping and Rotation Matrix}
For arbitrary coordinate systems, $\mathcal{A}$ and $\mathcal{B}$, with relative orientation $\Rot_{\mathcal{BA}} = (q_0, \check{\b{q}})$ the coordinates of a vector $\vec{\b{r}}$ can be mapped as:

\vspace{2pt}
\noindent
\colorbox{myRed}{\parbox[c][15pt]{\columnwidth-2\fboxsep}{
\begin{align}
	\Rot_{\mathcal{BA}}({}_{\mathcal{A}}\b{r}) = (2 q_0^2 - 1) {}_{\mathcal{A}}\b{r} + 2 q_0 \check{\b{q}}^\times \! {}_{\mathcal{A}}\b{r} + 2 \check{\b{q}} (\check{\b{q}}^T \! {}_{\mathcal{A}}\b{r}).
\end{align}
}}\vspace{2pt}
From this, we can directly derive the expression for the associated rotation matrix:

\vspace{2pt}
\noindent
\colorbox{myRed}{\parbox[c][15pt]{\columnwidth-2\fboxsep}{
\begin{align}
	\b{C}(\Rot_{\mathcal{BA}}) = (2 q_0^2 - 1) \b{I} + 2 q_0 \check{\b{q}}^\times + 2 \check{\b{q}} \check{\b{q}}^T.
\end{align}
}}\vspace{2pt}

\subsection{Concatenation}
The concatenation of two unit quaternions $\Rot_1 = (q_0, \check{\b{q}})$ and $\Rot_2 = (p_0, \check{\b{p}})$ is given by:

\vspace{2pt}
\noindent
\colorbox{myRed}{\parbox[c][15pt]{\columnwidth-2\fboxsep}{
\begin{align}
	\Rot_1 \circ \Rot_2 &= (q_0 p_0 - \check{\b{q}}^T \check{\b{p}}, q_0 \check{\b{p}} + p_0 \check{\b{q}} + \check{\b{q}} \times \check{\b{p}}).
\end{align}
}}\vspace{2pt}

\subsection{Exponential and Logarithm}
Given a $\rot \in \mathbb{R}^3$, the exponential map to a unit quaternion is given by:

\vspace{2pt}
\noindent
\colorbox{myRed}{\parbox[c][55pt]{\columnwidth-2\fboxsep}{
\begin{align}
	\exp(\rot) &= (q_0, \check{\b{q}}) = \left(\c(\|\rot\|/2) , \s(\|\rot\|/2) \frac{\rot}{\|\rot\|} \right) \\
	\exp(\rot) &\approx (1, \rot/2), \ \ (\|\rot\| \approx 0). \label{eq:quatExpLim}
\end{align}
}}\vspace{2pt}
The above small angle approximation is required to avoid numerical instabilities (typically for angles below \unit[$10^{-4}$]{rad}).
The corresponding logarithm is given by:

\vspace{2pt}
\noindent
\colorbox{myRed}{\parbox[c][38pt]{\columnwidth-2\fboxsep}{
\begin{align}
	\log(\Rot) &= 2 \, \textrm{atan2}(\|\check{\b{q}}\|,q_0) \frac{\check{\b{q}}}{\|\check{\b{q}}\|}, \\
	\log(\Rot) &\approx \textrm{sign}(q_0) \, \check{\b{q}}, \ \ (\|\check{\b{q}}\| \approx 0). \label{eq:quatLogLim}
\end{align}
}}\vspace{2pt}

\subsection{Consistency Tests}
The consistency of the implementation can be tested through the following unit tests:
\begin{center}
\begin{tabular}{l|l|l|l|l|l|}
& $\Rot()$ & $\b{C}$ & $\circ$ & $\exp$ & $\log$ \\\hline
$\b{C}(\Rot)\b{r} = \Rot(\b{r})$ & \cellcolor{myBlue} & \cellcolor{myBlue} & & & \\\hline
$(\Rot_1 \circ \Rot_2)(\b{r}) = \Rot_1 (\Rot_2(\b{r}))$ & \cellcolor{myBlue} & & \cellcolor{myBlue} & & \\\hline
$\b{C}(\exp(\rot)) = \b{C}(\rot)$ & & \cellcolor{myBlue} & & \cellcolor{myBlue} & \\\hline
$\Rot = \exp(\log(\Rot))$ & & & & \cellcolor{myBlue} & \cellcolor{myBlue} \\\hline
\end{tabular}
\end{center}
On the right-hand side the involved operators are listed.
The third test compares against Rodriguez' formula (\cref{eq:rod}).
Theoretically, these tests should be carried out for all possible values of $\Rot, \Rot_1, \Rot_2 \!\! \in \!\! SO(3), \ \b{r}, \b{\rot} \!\! \in \!\! \mathbb{R}^3$.
In practice, testing various samples, including very small angles, should be sufficient.

\section{Simple Modeling Example} \label{sec:modexa}
This section presents how to apply the above notation and convention to an actual system modeling task.
We want to estimate the position, velocity (expressed in $\mathcal{B}$ to simplify the Jacobians), and orientation of a robot using an IMU and a generic position and orientation sensor (pose sensor).
To avoid complicated modeling or specific knowledge about the motion model the IMU can be used to do a prediction of the state.
This is very common in visual-inertial state estimation e.g. \cite{bloesch2015robust}.
In the following, we first show how to use the IMU for predicting the state and then show the necessary steps to perform an update with the pose sensor.

\subsection{Continuous Time Description}
Let us assume we have an IMU driven dynamic system with inertial coordinate system $\mathcal{I}$ and IMU-fixed coordinate system $\mathcal{B}$ for which we wish to estimate the motion.
Considering additive biases, ${}_{\mathcal{B}}\b{b}_{f}$ and ${}_{\mathcal{B}}\b{b}_{\omega}$, and using continuous-time white noise processes, ${}_{\mathcal{B}}\b{n}_{f}$, ${}_{\mathcal{B}}\b{n}_{\omega}$, ${}_{\mathcal{B}}\b{n}_{bf}$, ${}_{\mathcal{B}}\b{n}_{b\omega}$, we can model the IMU measurements, ${}_{\mathcal{B}}\tilde{\b{f}}_{\mathcal{B}}$ and ${}_{\mathcal{B}}\tilde{\b{\omega}}_{\mathcal{B}}$, as:
\begin{align}
    {}_{\mathcal{B}}\tilde{\b{f}}_{\mathcal{B}} &= \Rot_{\mathcal{IB}}^{-1}({}_{\mathcal{I}}\b{a}_{\mathcal{B}} - {}_{\mathcal{I}}\b{g}) + {}_{\mathcal{B}}\b{b}_{f} + {}_{\mathcal{B}}\b{n}_{f}, \\
    {}_{\mathcal{B}}\tilde{\b{\omega}}_{\mathcal{B}} &= {}_{\mathcal{B}}\b{\omega}_{\mathcal{IB}} + {}_{\mathcal{B}}\b{b}_{\omega} + {}_{\mathcal{B}}\b{n}_{\omega}, \\
    {}_{\mathcal{B}}\dot{\b{b}}_{f} &= {}_{\mathcal{B}}\b{n}_{bf}, \\
    {}_{\mathcal{B}}\dot{\b{b}}_{\omega} &= {}_{\mathcal{B}}\b{n}_{b\omega},
\end{align}
where ${}_{\mathcal{I}}\b{g}$ is the gravity vector expressed in the inertial frame.
We add the IMU biases to the state $\b{x}$.
This gives the full state by
\begin{align}
      \b{x} &= \left(  {}_{\mathcal{I}}\b{r}_{\mathcal{IB}}, \,
      {}_{\mathcal{B}}\b{v}_{\mathcal{B}}, \,
      \Rot_{\mathcal{IB}}, \,
      {}_{\mathcal{B}}\b{b}_{f}, \,
      {}_{\mathcal{B}}\b{b}_{\omega}
      \right).
\end{align}
The resulting continuous-time equations of motion can be written as:
\begin{align}
    {}_{\mathcal{I}}\dot{\b{r}}_{\mathcal{IB}} &= \Rot_{\mathcal{IB}}({}_{\mathcal{B}}\b{v}_{\mathcal{B}} + {}_{\mathcal{B}}\b{n}_{v}), \\
    {}_{\mathcal{B}}\dot{\b{v}}_{\mathcal{B}} &= \nicefrac[]{d}{dt} \left(\Rot_{\mathcal{IB}}^{-1}({}_{\mathcal{I}}\b{v}_{\mathcal{B}})\right) \nonumber \\
    &= \Rot_{\mathcal{IB}}^{-1}({}_{\mathcal{I}}\dot{\b{v}}_{\mathcal{B}}) - \left(\Rot_{\mathcal{IB}}^{-1}({}_{\mathcal{I}}\b{v}_{\mathcal{B}})\right)^\times \b{C}(\Rot_\mathcal{IB})^{T} {}_{\mathcal{I}}\b{\omega}_{\mathcal{BI}} \nonumber \\
    &= \Rot_{\mathcal{IB}}^{-1}({}_{\mathcal{I}}\b{a}_{\mathcal{B}}) - {}_{\mathcal{B}}\b{v}_{\mathcal{B}}^\times {}_{\mathcal{B}}\b{\omega}_{\mathcal{BI}} \nonumber \\
    &= \Rot_{\mathcal{IB}}^{-1}({}_{\mathcal{I}}\b{g}) + {}_{\mathcal{B}}\b{f}_{\mathcal{IB}} - {}_{\mathcal{B}}\b{\omega}_{\mathcal{IB}}^\times {}_{\mathcal{B}}\b{v}_{\mathcal{B}}, \label{eq:c_vel_dyn}\\
    \dot{\Rot}_{\mathcal{IB}} &= -{}_{\mathcal{I}}\b{\omega}_{\mathcal{BI}} = \Rot_{\mathcal{IB}}({}_{\mathcal{B}}\b{\omega}_{\mathcal{IB}}), \\
    {}_{\mathcal{B}}\dot{\b{b}}_{f} &= {}_{\mathcal{B}}\b{n}_{bf}, \\
    {}_{\mathcal{B}}\dot{\b{b}}_{\omega} &= {}_{\mathcal{B}}\b{n}_{b\omega},
\end{align}
with the bias and noise corrected proper acceleration and angular velocity measurements
\begin{align}
    {}_{\mathcal{B}}\b{f}_{\mathcal{IB}} &= {}_{\mathcal{B}}\tilde{\b{f}}_{\mathcal{B}} - {}_{\mathcal{B}}\b{b}_{f} - {}_{\mathcal{B}}\b{n}_{f}, \\
    {}_{\mathcal{B}}\b{\omega}_{\mathcal{IB}} &= {}_{\mathcal{B}}\tilde{\b{\omega}}_{\mathcal{B}} - {}_{\mathcal{B}}\b{b}_{\omega} - {}_{\mathcal{B}}\b{n}_{\omega}.
\end{align}
To derive (\ref{eq:c_vel_dyn}) we used the product rule, followed by the chain rule and the identities (\ref{eq:dertime}),(\ref{eq:dercm}),(\ref{eq:derinv}).

\subsection{Euler-Forward Discretization}
One of the simplest and most commonly used discretization methods is Euler-Forward discretization.
Other discretization schemes can of course also be employed.
For a time increment $\Delta t$, Euler-Foward discretization of the above formulation yields (the next state is denoted by a bar, discretized noise by a hat):
\begin{align}
	{}_{\mathcal{I}}\bar{\b{r}}_{\mathcal{IB}} &= {}_{\mathcal{I}}\b{r}_{\mathcal{IB}} + \Delta t \ \Rot_{\mathcal{IB}}({}_{\mathcal{B}}\b{v}_{\mathcal{B}} + {}_{\mathcal{B}}\hat{\b{n}}_{v}), \\
    {}_{\mathcal{B}}\bar{\b{v}}_{\mathcal{B}} &= {}_{\mathcal{B}}\b{v}_{\mathcal{B}} + \Delta t \Big(\Rot_{\mathcal{IB}}^{-1}({}_{\mathcal{I}}\b{g}) + \b{f} - \b{\omega}^\times {}_{\mathcal{B}}\b{v}_{\mathcal{B}}\Big) \\
    \bar{\Rot}_{\mathcal{IB}} &=\Rot_{\mathcal{IB}} \boxplus (\Delta t \Rot_{\mathcal{IB}}(\b{\omega})) \nonumber \\
     &= \exp(\Rot_{\mathcal{IB}}(\Delta t \b{\omega})) \circ \Rot_{\mathcal{IB}} \nonumber \\
     &= \Rot_{\mathcal{IB}} \circ \exp(\Delta t \b{\omega}) \circ \Rot_{\mathcal{IB}}^{-1} \circ \Rot_{\mathcal{IB}} \nonumber \\
     &= \Rot_{\mathcal{IB}} \circ \exp(\Delta t \b{\omega}), \\
    {}_{\mathcal{B}}\bar{\b{b}}_{f} &= {}_{\mathcal{B}}\b{b}_{f} + \ \Delta t {}_{\mathcal{B}}\hat{\b{n}}_{bf}, \\
    {}_{\mathcal{B}}\bar{\b{b}}_{\omega} &= {}_{\mathcal{B}}\b{b}_{\omega} + \ \Delta t {}_{\mathcal{B}}\hat{\b{n}}_{b\omega},
\end{align}
with the discretized IMU measurements (bias and noise corrected) given by
\begin{align}
    \b{f} &= {}_{\mathcal{B}}\tilde{\b{f}}_{\mathcal{B}} - {}_{\mathcal{B}}\b{b}_{f} - {}_{\mathcal{B}}\hat{\b{n}}_{f}, \\
    \b{\omega} &= {}_{\mathcal{B}}\tilde{\b{\omega}}_{\mathcal{B}} - {}_{\mathcal{B}}\b{b}_{\omega} - {}_{\mathcal{B}}\hat{\b{n}}_{\omega}.
\end{align}
The noise is discretized such that, if $\b{R}_i$ is the noise density of the white noise process $\b{n}_i$, then the discrete Gaussian noise $\hat{\b{n}}_i$ is distributed with $\mathcal{N}(0,\b{R}_i/\Delta t)$.

\subsection{Differentiation}
Using the identities (\ref{eq:dertime})-(\ref{eq:derlogmap}) and applying the chain rule, the following Jacobians of the discrete process model can be derived ($\b{F}$ is w.r.t. the state, $\b{G}$ is w.r.t. the process noise):

\tiny
\begin{align}
	\b{F} & = \begin{bmatrix}
    	\b{I} & \Delta t \, \b{C}(\Rot_{\mathcal{IB}}) & -\Delta t \, \Rot_{\mathcal{IB}}({}_{\mathcal{B}}\b{v}_{\mathcal{B}})^\times & 0 & 0 \\
        0 & \b{I} - \Delta t \b{\omega}^\times & \Delta t \, \b{C}(\Rot_{\mathcal{IB}})^T({}_{\mathcal{I}}\b{g})^\times & -\Delta t \, \b{I} & -\Delta t \, {}_{\mathcal{B}}\b{v}_{\mathcal{B}}^\times \\
        0 & 0 & \b{I} & 0 & -\Delta t \b{C}(\Rot_{\mathcal{IB}}) \b{\Gamma}(\Delta t \b{\omega}) \\
        0 & 0 & 0 & \b{I} & 0 \\
        0 & 0 & 0 & 0 & \b{I}
    \end{bmatrix}, \nonumber \\
	\b{G} & = \begin{bmatrix}
    	\Delta t \, \b{C}(\Rot_{\mathcal{IB}}) & 0 & 0 & 0 & 0 \\
        0 & -\Delta t \b{I} & -\Delta t ({}_{\mathcal{B}}\b{v}_{\mathcal{B}})^\times & 0 & 0 \\
        0 & 0 & -\Delta t \b{C}(\Rot_{\mathcal{IB}}) \b{\Gamma}(\Delta t \b{\omega}) & 0 & 0 \\
        0 & 0 & 0 & \Delta t \b{I} & 0 \\
        0 & 0 & 0 & 0 & \Delta t \b{I}
    \end{bmatrix}, \nonumber
\end{align}
\normalsize

\subsection{Measurement}
For simplicity we assume a GPS position measurement ${}_{\mathcal{I}}\tilde{\b{r}}_{\mathcal{IB}}$ and an orientation measurement $\tilde{\Rot}_{\mathcal{IB}}$.
The measurement equations are given by
\begin{align}
    {}_{\mathcal{I}}\tilde{\b{r}}_{\mathcal{IB}} & = {}_{\mathcal{I}}\b{r}_{\mathcal{IB}} + {}_{\mathcal{I}}\hat{\b{n}}_{p}, \\
    \tilde{\Rot}_{\mathcal{IB}} & = \Rot_{\mathcal{IB}} \boxplus {}_{\mathcal{I}}\hat{\b{n}}_{\Rot} \\
     & = \exp({}_{\mathcal{I}}\hat{\b{n}}_{\Rot}) \circ \Rot_{\mathcal{IB}},
\end{align}
with the discrete Gaussian measurement noise vectors ${}_{\mathcal{I}}\hat{\b{n}}_{p}$ and ${}_{\mathcal{I}}\hat{\b{n}}_{\Rot}$.

Using the identities (\ref{eq:dercon1}),(\ref{eq:dercon2}),(\ref{eq:derexp}) and because the expectation of the orientation measurement noise is zero, the following Jacobians can be derived ($\b{H}$ is w.r.t. the state, $\b{J}$ is w.r.t. the update noise):
\begin{align}
	\b{H} & = \begin{bmatrix}
    	\b{I} & 0 & 0 & 0 & 0 \\
        0 & 0 & \b{I} & 0 & 0
    \end{bmatrix}, \nonumber \\
	\b{J} & = \begin{bmatrix}
    	\b{I} & 0 \\
        0 & \b{I}
    \end{bmatrix}, \nonumber
\end{align}

\subsection{Hints for the EKF Implementation}
Now that we have derived all the required parts, the well known EKF equations can be used to estimate the state.
The only difference to the standard EKF is that we need to use the $\boxminus$ operator for the innovation residual and the $\boxplus$ operator for updating the state estimate instead of normal addition and subtraction.

\section{Conclusion}
This document derived and summarized the main identities related to 3D orientations in robotics and other engineering fields.
In particular it discussed a more abstract but convention-less notion of 3D orientations, the boxplus and boxminus operators, as well as the concept of differentials.
Various differentials involving 3D orientations are derived, which can be used to compute the Jacobians of more complex models by applying the chain rule.
A simple modeling example shows how to apply the introduced concepts.

\appendices
\section{Derivatives Involving Orientations}\label{app:der}
\subsubsection{Time Derivative of Orientation}
Here we need the kinematic concept of angular velocities.
We assume the existence of an inertial observer $\mathcal{I}$ which observes the motion, over a duration $\epsilon$, of a moving coordinate system $\mathcal{B}(t)$.
We use the following definition of angular velocities (the negative sign is required so that the angular velocity corresponds to the \emph{active} rotation which is measured by typical IMU devices):
\begin{align}
	{}_{\mathcal{B}(t)}\b{\omega}_{\mathcal{I}\mathcal{B}(t)} := -\lim_{\epsilon \rightarrow 0} \frac{{}_{\mathcal{B}(t)}\rot_{\mathcal{B}(t+\epsilon)\mathcal{B}(t)}}{\epsilon} \label{eq:angvel}
\end{align}
Additionally we require the limit (based on the limits (\ref{eq:quatExpLim}),(\ref{eq:quatLogLim})):
\begin{align}
	\lim_{\epsilon \rightarrow 0} \frac{\log(\exp(\epsilon \rot_1) \circ \exp(\epsilon \rot_2))}{\epsilon} = \rot_1 + \rot_2. \label{eq:logadd}
\end{align}
With this we can derive the derivative of an orientation $\Rot_{\mathcal{B}(t)\mathcal{A}(t)}$ w.r.t. time $t$ (used identities: (\ref{eq:der1}),(\ref{eq:boxminus}),(\ref{eq:rotvec}),(\ref{eq:adjoint}),(\ref{eq:angvel}),(\ref{eq:logadd})):
\begin{align}
	\frac{\partial}{\partial t} \Rot_{\mathcal{B}(t)\mathcal{A}(t)} =& \lim_{\epsilon \rightarrow 0} \frac{\Rot_{\mathcal{B}(t+\epsilon)\mathcal{A}(t+\epsilon)} \boxminus \Rot_{\mathcal{B}(t)\mathcal{A}(t)}}{\epsilon} \nonumber \\
	=& \lim_{\epsilon \rightarrow 0} \frac{1}{\epsilon} \Big((\Rot_{\mathcal{B}(t+\epsilon)\mathcal{B}(t)} \circ \Rot_{\mathcal{B}(t)\mathcal{A}(t)} \circ \Rot_{\mathcal{A}(t)\mathcal{A}(t+\epsilon)}) \nonumber \\
	& \ \ \ \ \ \ \ \ \ \boxminus \Rot_{\mathcal{B}(t)\mathcal{A}(t)} \Big) \nonumber \\
	=& \lim_{\epsilon \rightarrow 0} \frac{1}{\epsilon} \Big(\log(\Rot_{\mathcal{B}(t+\epsilon)\mathcal{B}(t)} \circ \Rot_{\mathcal{B}(t)\mathcal{A}(t)} \nonumber \\
	& \ \ \ \ \ \ \ \ \ \circ \Rot_{\mathcal{A}(t)\mathcal{A}(t+\epsilon)} \circ \Rot_{\mathcal{B}(t)\mathcal{A}(t)}^{-1}) \Big) \nonumber \\
	=& \lim_{\epsilon \rightarrow 0} \frac{1}{\epsilon} \Big(\log(\exp({}_{\mathcal{B}(t)}\rot_{\mathcal{B}(t+\epsilon)\mathcal{B}(t)}) \circ \Rot_{\mathcal{B}(t)\mathcal{A}(t)} \nonumber \\
	& \ \ \ \ \ \ \ \ \ \circ \exp({}_{\mathcal{A}(t)}\rot_{\mathcal{A}(t)\mathcal{A}(t+\epsilon)}) \circ \Rot_{\mathcal{B}(t)\mathcal{A}(t)}^{-1})\Big) \nonumber \\
	=& \lim_{\epsilon \rightarrow 0} \frac{1}{\epsilon} \Big(\log(\exp({}_{\mathcal{B}(t)}\rot_{\mathcal{B}(t+\epsilon)\mathcal{B}(t)}) \nonumber \\
	& \ \ \ \ \ \ \ \ \ \circ \exp({}_{\mathcal{B}(t)}\rot_{\mathcal{A}(t)\mathcal{A}(t+\epsilon)}))\Big) \nonumber \\
	=& \lim_{\epsilon \rightarrow 0} \frac{1}{\epsilon} \Big(\log(\exp(-\epsilon \, {}_{\mathcal{B}(t)}\b{\omega}_{\mathcal{I}\mathcal{B}(t)}) \nonumber \\
	& \ \ \ \ \ \ \ \ \ \circ \exp(\epsilon \, {}_{\mathcal{B}(t)}\b{\omega}_{\mathcal{I}\mathcal{A}(t)}))\Big) \nonumber \\
	=& -{}_{\mathcal{B}(t)}\b{\omega}_{\mathcal{I}\mathcal{B}(t)} + {}_{\mathcal{B}(t)}\b{\omega}_{\mathcal{I}\mathcal{A}(t)} \nonumber \\
	=& -{}_{\mathcal{B}(t)}\b{\omega}_{\mathcal{A}(t)\mathcal{B}(t)}
\end{align}
\subsubsection{Derivative of Inverse}
Here we derive the derivative of the inverse of an orientation (used identities: (\ref{eq:der1}),(\ref{eq:der2}),(\ref{eq:boxminus}),(\ref{eq:boxplus}),(\ref{eq:adjoint})):
\begin{align}
	\left[\frac{\partial}{\partial \Rot} \Rot^{-1}\right]_i =& \lim_{\epsilon \rightarrow 0} \frac{(\Rot \boxplus \b{e}_i \epsilon )^{-1} \boxminus \Rot^{-1}}{\epsilon} \nonumber \\
	=& \lim_{\epsilon \rightarrow 0} \frac{\log(\Rot^{-1} \circ \exp(-\b{e}_i \epsilon) \circ \Rot)}{\epsilon} \nonumber \\
	=& \lim_{\epsilon \rightarrow 0} \frac{\log(\exp(-\Rot^{-1}(\b{e}_i) \epsilon)}{\epsilon} \nonumber \\
	=& -\Rot^{-1}(\b{e}_i) =  -\b{C}(\Rot)^T \b{e}_i. \nonumber \\
	\frac{\partial}{\partial \Rot} \Rot^{-1} =& -\b{C}(\Rot)^T.
\end{align}

\subsubsection{Derivative of Coordinate Map}
The map of an orientation applied to a coordinate tuple can be differentiated w.r.t. the orientation itself.
This yields (used identities: (\ref{eq:der2}),(\ref{eq:boxplus}),(\ref{eq:condef}),(\ref{eq:rodsmall})):
\begin{align}
	\left[\frac{\partial}{\partial \Rot} \Rot(\b{r})\right]_i =& \lim_{\epsilon \rightarrow 0} \frac{(\Rot \boxplus \b{e}_i \epsilon )(\b{r}) - \Rot(\b{r})}{\epsilon} \nonumber \\
	=& \lim_{\epsilon \rightarrow 0} \frac{\b{C}(\b{e}_i \epsilon)\b{C}(\Rot)\b{r} - \b{C}(\Rot)\b{r}}{\epsilon} \nonumber \\
	=& \lim_{\epsilon \rightarrow 0} \frac{(\b{I}+\b{e}_i^\times \epsilon)\b{C}(\Rot)\b{r} - \b{C}(\Rot)\b{r}}{\epsilon} \nonumber \\
	=& \lim_{\epsilon \rightarrow 0} \frac{\b{e}_i^\times \epsilon \b{C}(\Rot)\b{r}}{\epsilon} \nonumber \\
	=& -(\b{C}(\Rot)\b{r})^\times \b{e}_i. \nonumber \\
	\frac{\partial}{\partial \Rot} \Rot(\b{r}) =& -(\b{C}(\Rot)\b{r})^\times.
\end{align}

\subsubsection{Concatenation - Left}
The concatenation of two orientations can be differentiated w.r.t. the involved orientations.
We first derive the derivative w.r.t. the left orientation (used identities: (\ref{eq:der1}),(\ref{eq:der2}),(\ref{eq:boxminus}),(\ref{eq:boxplus})):
\begin{align}
	\left[\frac{\partial}{\partial \Rot_1} \Rot_1 \circ \Rot_2 \right]_i =& \lim_{\epsilon \rightarrow 0} \frac{((\Rot_1 \boxplus \b{e}_i \epsilon) \circ \Rot_2) \boxminus (\Rot_1 \circ \Rot_2)}{\epsilon} \nonumber \\
	=& \lim_{\epsilon \rightarrow 0} \frac{\log(\exp(\b{e}_i \epsilon) \circ \Rot_1 \circ \Rot_2 \circ \Rot_2^{-1} \circ \Rot_1^{-1})}{\epsilon} \nonumber \\
	=& \ \b{e}_i. \nonumber \\
	\frac{\partial}{\partial \Rot_1} \Rot_1 \circ \Rot_2 =& \ \b{I}.
\end{align}

\subsubsection{Concatenation - Right}
The derivative of the concatenation w.r.t. the right orientation yields (used identities: (\ref{eq:der1}),(\ref{eq:der2}),(\ref{eq:boxminus}),(\ref{eq:boxplus}),(\ref{eq:adjoint})):
\begin{align}
	\left[\frac{\partial}{\partial \Rot_2} \Rot_1 \circ \Rot_2 \right]_i =& \lim_{\epsilon \rightarrow 0} \frac{(\Rot_1 \circ (\Rot_2 \boxplus \b{e}_i \epsilon)) \boxminus (\Rot_1 \circ \Rot_2)}{\epsilon} \nonumber \\
	=& \lim_{\epsilon \rightarrow 0} \frac{\log( \Rot_1 \circ \exp(\b{e}_i \epsilon) \circ \Rot_2 \circ \Rot_2^{-1} \circ \Rot_1^{-1})}{\epsilon} \nonumber \\
	=& \lim_{\epsilon \rightarrow 0} \frac{\log( \exp(\Rot_1(\b{e}_i) \epsilon))}{\epsilon} \nonumber \\
	=& \ \Rot_1(\b{e}_i) = \b{C}(\Rot_1) \b{e}_i. \nonumber \\
	\frac{\partial}{\partial \Rot_2} \Rot_1 \circ \Rot_2 =& \ \b{C}(\Rot_1).
\end{align}

\subsubsection{Exponential Derivative}
Define:
\begin{align}
	\b{\Gamma}(\rot) &:= \partial/\partial \rot \left(\exp(\rot)\right). \label{eq:gammadef}
\end{align}
Differentiate the adjoint related identity using the chain rule and product rule (identities (\ref{eq:gammadef}),(\ref{eq:dercm}) for left side, identities (\ref{eq:dercon1}),(\ref{eq:dercon2}),(\ref{eq:derinv}) for right side):
\begin{align}
	\partial/\partial \Rot \Big[ \exp(\Rot(\rot)) &= \Rot \circ \exp(\rot) \circ \Rot^{-1} \Big], \\
	-\b{\Gamma}(\Rot(\rot)) \Rot(\rot)^\times &= \b{I} - \b{C}(\Rot) \b{C}(\rot) \b{C}(\Rot)^T.
\end{align}
Set $\Rot$ to identity:
\begin{align}
	\b{\Gamma}(\rot) \rot^\times = \b{C}(\rot) - \b{I}. \label{eq:expder1}
\end{align}
Now consider the map $f(x) = \exp(x\rot)$ for some arbitrary $\rot \in \mathbb{R}^3$.
The chain rule yields $f'(x) = \b{\Gamma}(x\rot) \rot$.
Alternatively, it can be differentiated using the limit (\ref{eq:der1}) (used identities: (\ref{eq:expCon1}),(\ref{eq:boxminus})):
\begin{align}
	f'(x) &= \lim_{\epsilon \rightarrow 0} \frac{\exp((x+\epsilon)\rot) \boxminus \exp(x\rot)}{\epsilon} \nonumber \\
    &= \lim_{\epsilon \rightarrow 0} \frac{\log(\exp(\epsilon\rot) \circ \exp(x\rot) \circ \exp(x\rot)^{-1})}{\epsilon} \nonumber \\
    &= \rot.
\end{align}
Compare both derivatives at $x=1$:
\begin{align}
	\b{\Gamma}(\rot) \rot &= \rot.
\end{align}
This can be combined with \cref{eq:expder1} in order to obtain the following matrix equation:
\begin{align}
	\b{\Gamma}(\rot) \begin{bmatrix}\rot^\times & \rot\end{bmatrix} &= \begin{bmatrix} \b{C}(\rot) - \b{I} & \rot\end{bmatrix}.
\end{align}
Right multiply with $\begin{bmatrix}\rot^\times & \rot\end{bmatrix}^T$ and simplify:
\begin{align}
	\b{\Gamma}(\rot) (-\rot^{\times^2} + \rot\rot^T) &= (\b{I} - \b{C}(\rot))\rot^\times +  \rot\rot^T, \\
	\b{\Gamma}(\rot) \|\rot\|^2 &= (\b{I} - \b{C}(\rot))\rot^\times +  \rot\rot^T, \\
	\b{\Gamma}(\rot) &= \frac{(\b{I} - \b{C}(\rot))\rot^\times +  \rot\rot^T}{\|\rot\|^2}.
\end{align}
If substituting $\b{C}(\rot)$ we obtain \cref{eq:gamma}.

\section{Other Proofs}
\subsection{Rodriguez' Formula} \label{ap:rodfor}

From \cref{eq:exppCon1,eq:exppCon2,eq:condef} we obtain the following properties for $\b{C}(\rot) = \b{C}(\exp(\rot))$, $\forall t \in \mathbb{R}, \, \rot,\b{v} \in \mathbb{R}^3$:
\begin{align}
  \b{C}((t+s) \rot) &= \b{C}(t \rot) \b{C}(s \rot) \\
  \nicefrac[]{d}{dt} \left(\b{C}(t \rot)(\b{v})\right)|_{t=0} &= \rot \times \b{v}
\end{align}
For a given $\rot$ we define the curve $\b{C}_{\rot}(t) := \b{C}(t \rot)$.
Using a change of coordinate $t = s + r$, we can extend the range of the differential identity $\forall t \in \mathbb{R}, \, \b{v} \in \mathbb{R}^3$:
\begin{align}
  \nicefrac[]{d}{dt} \left(\b{C}(t \rot)\b{v}\right) &= \nicefrac[]{d}{ds} \left(\b{C}(s \rot)\b{C}(r \rot)\b{v}\right)|_{s=0, r = t} \\
  &= \rot^\times \b{C}(t \rot)\b{v}
\end{align}
Thus, we obtain the following matrix differential equation:
\begin{align}
  \nicefrac[]{d}{dt} \left(\b{C}_{\rot}(t)\right) = \rot^\times \b{C}_{\rot}(t),
\end{align}
which has the matrix exponential solution
\begin{align}
  \b{C}_{\rot}(t) = e^{t \rot^\times}.
\end{align}
Since this is valid for arbitrary $\rot$, we obtain:
\begin{align}
  \b{C}(\rot) = e^{\rot^\times},
\end{align}
which can be shown to be the same as \cref{eq:rod} using series expansions.

\subsection{Concatenation and Exponential -- Adjoint Related} \label{ap:adjoint}
We want to prove the following identity:
\begin{align}
	\exp(\Rot(\rot)) &= \Rot \circ \exp(\rot) \circ \Rot^{-1}.
\end{align}
Since we know that $\exp$ is unique it is sufficient to show that the right hand side is indeed the exponential of $\Rot(\rot)$ and thus check the defining properties.
First we verify \cref{eq:exppCon1}:
\begin{align}
	&\exp((t+s) \Rot(\rot)) \\
	&= \Rot \circ \exp((t+s) \rot) \circ \Rot^{-1} \\
	&= \Rot \circ \exp(t \rot) \circ \exp(s \rot) \circ \Rot^{-1} \\
	&= \Rot \circ \exp(t \rot) \circ \Rot^{-1} \circ \Rot \circ \exp(s \rot) \circ \Rot^{-1} \\
	&= \exp(t \Rot(\rot)) \circ \exp(s \Rot(\rot)).
\end{align}
\Cref{eq:exppCon2} poses a requirement on the derivative which can also be verified:
\begin{align}
  &\nicefrac[]{d}{dt} \left(\exp(t \Rot(\rot))(\b{v})\right)|_{t=0} \\
	&= \nicefrac[]{d}{dt} \left(\b{C}(\Rot) \left(\exp(t \rot)(\b{C}(\Rot)^T \b{v})\right)\right)|_{t=0} \\
	&= \b{C}(\Rot) \rot^\times \b{C}(\Rot)^T \b{v} \\
	&= (\b{C}(\Rot) \rot)^\times \b{v} = \Rot(\rot) \times \b{v}.
\end{align}
Since $\Rot \circ \exp(\rot) \circ \Rot^{-1}$ fulfills both uniquely defining properties of the exponential it is indeed equivalent to $\exp(\Rot(\rot))$.

\bibliographystyle{IEEEtran}
\bibliography{ref.bib}
\end{document}